\documentclass{article}
\usepackage{spconf,amsmath,mathtools,graphicx}
\usepackage{amssymb}

\usepackage{microtype}
\usepackage{amssymb}
\usepackage{amsthm} 
\usepackage{subfigure}
\usepackage{mathtools}
\usepackage{xcolor} 
\usepackage{tikz} 
\usepackage{sidecap} 
\usepackage{hyperref} 
\usepackage{cite}
\usepackage{wrapfig}
\usepackage{adjustbox}

\newcommand{\diag}{\operatorname{diag}}
\title{Geometric Scattering Attention Networks}
\name{Yimeng Min$^{1,3,\star}$ \qquad Frederik Wenkel$^{2,3,\star}$ \qquad Guy Wolf$^{\,2,3}$ \thanks{$^\star$ Equal contribution; student authors; order determined alphabetically. Correspondence to \texttt{guy.wolf@umontreal.ca}$\,$. This work was partially funded by IVADO Professor startup \& operational funds, IVADO Fundamental Research Proj.\ grant PRF-2019-3583139727, and NIH grant R01GM135929. The content provided here is solely the responsibility of the authors and does not necessarily represent the official views of the funding agencies.}}
\address{Universit\'{e} de Montr\'{e}al, $^{1}$Department of Computer Science \& Operational Research \\ $^{2}$Department of Mathematics \& Statistics; $^{3}$Mila -- Quebec AI Institute, Montreal, QC, Canada}
\begin{document}
\maketitle
\begin{abstract}
Geometric scattering has recently gained recognition in graph representation learning, and recent work has shown that integrating scattering features in graph convolution networks (GCNs) can alleviate the typical oversmoothing of features in node representation learning. However, scattering often relies on handcrafted design, requiring careful selection of frequency bands via a cascade of wavelet transforms, as well as an effective weight sharing scheme to combine low- and band-pass information. Here, we introduce a new attention-based architecture to produce adaptive task-driven node representations by implicitly learning node-wise weights for combining multiple scattering and GCN channels in the network. We show the resulting geometric scattering attention network (GSAN) outperforms previous networks in semi-supervised node classification, while also enabling a spectral study of extracted information by examining node-wise attention weights.
\end{abstract}
\nopagebreak
\section{Introduction}
Convolutional neural networks (CNNs) have shown great success on a range of tasks, including image classification, machine translation and speech recognition. By optimizing local filters in neural network-based architectures, models are able to learn expressive representations and thus perform well on regular Euclidean data. Based on CNNs, graph neural networks (GNNs)~\cite{bronstein2017geometric, kipf2016semi, velivckovic2018graph, hamilton2017inductive} show promising results on non-Euclidean data for tasks such as molecule modelling or node classification. The generalization from regular grids to irregular domains is usually implemented using the spectral graph theory framework~\cite{bruna2013spectral, defferrard2016convolutional}. While several approaches exist to implement such filters, most popular GNNs (cf.\ \cite{kipf2016semi,velivckovic2018graph,hamilton2017inductive}) tend to implement message passing operations that aggregate neighbourhood information using a one-step neighbourhood-localized filter, which corresponds to the lowest frequency of the graph Laplacian.

Previous studies suggest the convolutional operation of such local-smoothing, which can be interpreted as low-pass filtering on the feature vectors \cite{nt2019revisiting, dwivedi2020benchmarking}, forces a smooth embedding of neighbouring nodes, leading to information loss from message passing and severely degrading the performance \cite{li2018deeper,oono2019graph}. To assign different weights to the neighbouring nodes, graph attention networks (GATs) use attention layers to learn the adaptive weights across the edges, resulting in a leap in model capacity~\cite{velivckovic2018graph}. Such attention layers additionally increase the model interpretability. However, the resulting networks still rely on averaging neighboring node features for similarity computations, rather than leveraging more complex patterns. 

In order to capture higher-order regularity on graphs, geometric scattering networks were recently introduced~\cite{gao2019geometric, gama2019diffusion, gama2019stability, zou2020graph}. These generalize the Euclidean scattering transform~\cite{mallat2012group, bruna:invariantScatConvNet2013, anden:deepScatSpectrum2014} to the graph domain and leverage graph wavelets to extract effective and efficient graph representations. In~\cite{min2020scattering}, a hybrid scattering graph convolutional network (Sc-GCN) is proposed in order to tackle oversmoothing in traditional GCNs \cite{kipf2016semi}. Geometric scattering together with GCN-based filters are used to apply both band-pass and low-pass filters to the graph signal.
We note that some non-hybrid approaches have been proposed to learn band-pass filters via their spectral coefficients, but their advantages over smoothing (or low-pass) based architectures are inconclusive on node level tasks (see, e.g., studies in~\cite{bianchi2021graph}). Furthermore, as shown in \cite{min2020scattering}, the hybrid Sc-GCN approach significantly outperforms such approaches (in particular~\cite{defferrard2016convolutional}) on several benchmarks. However, even though Sc-GCN achieves good performance on a range of node level classification tasks, it requires the selection of a task-appropriate configuration of the network and its scattering wavelet composition to carefully balance low-pass and band-pass information. 

Here, we introduce a geometric scattering attention network (GSAN) that combines the hybrid Sc-GCN approach with a node-wise attention mechanism to automatically adapt its filter (or \emph{channel}) composition, thus simplifying its architecture tuning. We evaluate our proposed approach on a variety of semi-supervised node classification benchmarks, demonstrating its performance improvement over previous GNNs. Analyzing the node-wise distributions of attention weights further enables a deeper understanding of the network mechanics by relating the node-level task-dependent information (label) with the corresponding feature selection.
\section{Preliminaries}\label{sec:preliminaries}
We consider a weighted graph $G = (V,E,w)$ with nodes $V\coloneqq \{v_1,\dots,v_n\}$ and (undirected) edges $E\subset \{\{v_i, v_j\}\in V\times V , i\neq j\}$. The function $w : E \to (0,\infty)$ assigns positive weights to the graph edges, which we aggregate in the \textit{adjacency matrix} $\boldsymbol{W}\in\mathbb{R}^{n\times n}$ via
\[
    \boldsymbol{W}[v_i,v_j] \coloneqq
    \begin{cases} 
    w(v_i,v_j) & \text{if } \{v_i,v_j\}\in E, \\
    0 & \text{otherwise.}
    \end{cases}
\]
Each node $v_i \in V$ possesses a feature vector $\boldsymbol{x}_i \in \mathbb{R}^{d_0}$. These are aggregated in the feature matrix $\boldsymbol{X}\in\mathbb{R}^{n\times d_0}$. We further define the \textit{degree matrix} $\boldsymbol{D}\in\mathbb{R}^{n\times n}$, defined by $\boldsymbol{D}\coloneqq\boldsymbol{D}(\boldsymbol{W})\coloneqq \diag(d_1,\dots, d_n)$ with $d_i\coloneqq \deg(v_i)\coloneqq \sum_{j=1}^n \boldsymbol{W}[v_i,v_j]$ being the \textit{degree} of the node $v_i$ and $\diag(.)$ a diagonal matrix parameterized by the diagonal elements. The GNN methods discussed in the following yield layer-wise node representations, compactly written as $\boldsymbol{H}^{(\ell)}\in\mathbb{R}^{n\times d_\ell}$ for the $\ell^{th}$ layer with $\boldsymbol{H}^{(0)}\coloneqq\boldsymbol{X}$.

\subsection{Graph Convolutional Networks}
A very popular method introduced in \cite{kipf2016semi} connects the local node-based information in $\boldsymbol{X}$ with the intrinsic data-geometry encoded by $\boldsymbol{W}$. This is realized by filtering the node features with the layer-wise update rule
\begin{equation}\label{eq_gcn}
    \boldsymbol{H}^{(\ell)}  = \sigma(\boldsymbol{A} \boldsymbol{H}^{(\ell-1)} \boldsymbol{\Theta}^{(\ell)}).
\end{equation}
The matrix multiplication with
$$
\boldsymbol{A}\coloneqq (\boldsymbol{D}+\boldsymbol{I}_n)^{-1/2} \left(\boldsymbol{W}+\boldsymbol{I}_n\right) (\boldsymbol{D}+\boldsymbol{I}_n)^{-1/2}
$$
constitutes the filtering operation, while the multiplication with $\boldsymbol{\Theta}^{(\ell)}$ can be seen as a fully connected layer applied to the node features. Lastly an elementwise nonlinerity $\sigma(.)$ is applied.

This method is subject to the so-called oversmooting problem \cite{li2018deeper}, which causes the node features to be smoothed out, the more GCN layers are iterated. In signal processing terminology, the update rule can be interpreted as a low-pass filtering operation \cite{min2020scattering} so that the model cannot access a significant potion of the information considered in the frequency domain.


\subsection{Geometric Scattering}\label{sec:scattering}
Recently, geometric scattering was introduced to incorporate band-pass filters in GNNs~\cite{gao2019geometric,min2020scattering}, inspired by the utilization of scattering features in the analysis of images~\cite{bruna:invariantScatConvNet2013,mallat:rotoScat2013} and audio signals~\cite{anden:deepScatSpectrum2014,lostanlen:scatPitchSpiral2015}. Cascades of wavelets can often recover high-frequency information and geometric scattering exhibits an analogous property on graph domains.

Geometric scattering is based on the lazy random walk matrix
$
    \boldsymbol{P} \coloneqq \frac{1}{2} \big( \boldsymbol{I}_n + \boldsymbol{W} \boldsymbol{D}^{-1} \big),
$
which is used to construct diffusion wavelet matrices $\boldsymbol{\Psi}_k\in\mathbb{R}^{n\times n}$~\cite{coifman:diffWavelets2006} of order $k\in\mathbb{N}_0$,
\begin{equation}\label{eq_wavelet matrix}
    \begin{cases}
    \boldsymbol{\Psi}_0 \coloneqq \boldsymbol{I}_n - \boldsymbol{P}, \\
    
    \boldsymbol{\Psi}_k \coloneqq \boldsymbol{P}^{2^{k-1}} - \boldsymbol{P}^{2^k}, \quad k\geq 1.
\end{cases}
\end{equation}
For node features $\boldsymbol{H}$, the scattering features are calculated as
$$
     \boldsymbol{U}_p \boldsymbol{H} \coloneqq \boldsymbol{\Psi}_{k_m} \vert \boldsymbol{\Psi}_{k_{m-1}} \dots \vert \boldsymbol{\Psi}_{k_2} \vert \boldsymbol{\Psi}_{k_1}\boldsymbol{H}\vert \vert \dots \vert,
$$ 
with $p \coloneqq (k_1, \dots, k_m)\in \cup_{m \in \mathbb{N}} \mathbb{N}_0^{m}$ parameterizing the sequence of wavelets, which are separated by elementwise absolute value operations.
The layer-wise update rule has the form
\begin{equation}\label{eq_sct}
    \boldsymbol{H}^{(\ell)} \coloneqq \sigma \left( \boldsymbol{U}_{p} \boldsymbol{H}^{(\ell-1)} \boldsymbol{\Theta}^{(\ell)}\right).
\end{equation}

In \cite{min2020scattering}, a hybrid architecture (referred to as Sc-GCN here) is proposed in order to combine the benefits of both GCN and scattering filters. Therefore, \textit{network channels} $\big\{ \boldsymbol{H}_i^{(\ell)} \big\}_{i=1}^m$, each coming from either GCN (Eq.~\ref{eq_gcn}) or scattering (Eq.~\ref{eq_sct}), are concatenated (horizontally), constituting the \textit{hybrid layer}
$$
    \boldsymbol{H}^{(\ell)}\coloneqq\left[ \boldsymbol{H}_1^{(\ell)} \mathbin\Vert \dots \mathbin\Vert \boldsymbol{H}_m^{(\ell)} \right].
$$


\subsection{Graph Residual Convolution}\label{subsec_residual convolution}
This architecture component from \cite{min2020scattering} constitutes an adjustable low-pass filter, parameterized by the matrix
$$
    \boldsymbol{A}_{res}(\alpha) = \frac{1}{\alpha+1} (\boldsymbol{I}_n+\alpha \boldsymbol{W} \boldsymbol{D}^{-1}),
$$
which is usually applied to $\boldsymbol{H}^{(\ell)}$ followed by a fully connected layer (without nonlinearity). It filters the hybrid layer output for high-frequency noise, which can occur as a result of scattering features.

\subsection{Graph Attention Networks}
Another popular approach for node classification tasks was introduced in~\cite{velivckovic2018graph}, where at any node $v_i$, an attention mechanism attends over the aggregation of node features from the node neighborhood $\mathcal{N}_i$. The aggregation coefficients are learned via
\begin{equation*}
    \alpha_{ij} = \frac{\text{exp(LeakyReLU}({\boldsymbol{a}^T\left[\boldsymbol{\Theta h}_i\mathbin\Vert \boldsymbol{\Theta h}_j\right]})}{\sum_{v_k \in \mathcal{N}_i} \text{exp(LeakyReLU}(\boldsymbol{a}^T\left[\boldsymbol{\Theta h}_i\mathbin\Vert \boldsymbol{\Theta h}_k\right])}.
\end{equation*}
where $\boldsymbol{h}_i \in \mathbb{R}^d$ is the feature of node $i$, $\boldsymbol{\Theta} \in \mathbb{R}^{d' \times d}$ is the weight matrix and $\boldsymbol{a} \in \mathbb{R}^{2d'}$ is the attention vector. The output feature is $\boldsymbol{h}'_i = \sigma(\sum_{j \in \mathcal{N}_i}  \alpha_{ij} \boldsymbol{\Theta}\boldsymbol{h}_j)$.
For more expressivity, multi-head attention is used to generate concatenated features,
\begin{equation*}\textstyle
    \boldsymbol{h}'_i = \mathbin\Vert_{k=1}^K \sigma\left({\sum_{j\in \mathcal{N}_i}} \alpha_{ij}^k \boldsymbol{\Theta}_k \boldsymbol{h}_j\right),
\end{equation*}
where $K$ is the number of attention heads.

\section{Scattering attention layer}
\label{sec:attentionlayer}
Inspired by the recent work in Sec.~\ref{sec:preliminaries}, we introduce an attention framework to combine multiple channels corresponding to GCN and scattering filters while adaptively assigning different weights to them based on filtered node features. While our full network uses multi-head attention, we focus here on the processing performed independently by each attention head, deferring the multi-head configuration details to the general discussion of network architecture in the next section.

For every attention head, we first linearly transform the feature matrix $\boldsymbol{H}^{(\ell-1)}$ with a matrix $\boldsymbol{\Theta}^{(\ell)}$, setting the transformed feature matrix to be $\boldsymbol{\bar H}^\ell =  \boldsymbol{H}^{(\ell-1)}\boldsymbol{\Theta}^{(\ell)}$.
Then, based on the scattering GCN approach (Sec.~\ref{sec:scattering} and~\cite{min2020scattering}), a set of $C_{gcn}$ GCN channels and $C_{sct}$ scattering channels are calculated,
\begin{equation}\label{sct_layer}
\begin{cases}
\left.\begin{aligned}
    &\boldsymbol{\bar H}_{gcn,1}^{(\ell)} = \boldsymbol{A} \boldsymbol{\bar H}^{(\ell)}  \\
    & \; \vdots \\
    &\boldsymbol{\bar H}_{gcn,C_{gcn}}^{(\ell)} = \boldsymbol{A}^{C_{gcn}} \boldsymbol{\bar H}^{(\ell)} \\
    \end{aligned}  \right\} \text{ GCN channels,} \\
    \ \\
\left.\begin{aligned} 
    & \boldsymbol{\bar H}_{sct,1}^{(\ell)} = \vert \boldsymbol{U}_{p_1} \boldsymbol{\bar H}^{(\ell)} \vert ^q\\
    & \; \vdots \\
    & \boldsymbol{\bar H}_{sct,C_{sct}}^{(\ell)} = \vert \boldsymbol{U}_{p_{C_{sct}}}\boldsymbol{\bar H}^{(\ell)} \vert^q
    \end{aligned}\right\} \text{ scattering channels.}
 \end{cases}
\end{equation}
The channels $\boldsymbol{\bar H}_{gcn,i}^{(\ell)}$ perform low-pass operations with different spatial support, aggregating information from 1,$\dots$,$C_{gcn}$-step neighborhoods, respectively, while $\boldsymbol{\bar H}_{sct,k}^{(\ell)}$, defined according to Eq.~\ref{eq_sct}, enables band-pass filtering of graph signals.

Next, we compute attention coefficients that will be used in a shared attention layer to combine the filtered channels. In order to compute these node-wise attention coefficient for each channel, we first compute
\begin{align*}
     \boldsymbol{e}_{gcn,i}^{(\ell)} &= \operatorname{LeakyReLU} \left(\left[ \boldsymbol{\bar H}^{(\ell)} \Vert \boldsymbol{\bar H}_{gcn,i}^{(\ell)} \right] \boldsymbol{a} \right), 
\end{align*}
with analogous $\boldsymbol{e}_{sct,i}^{(\ell)}$ and $\boldsymbol{a} \in \mathbb{R}^{2 d_{\ell}}$ being a shared attention vector across all channels. 
We interpret $\boldsymbol{e}_{gcn,j}^{(\ell)}, \boldsymbol{e}_{sct,k}^{(\ell)} \in \mathbb{R}^n$ as score vectors indicating the importance of each channel.

Finally, the attention scores are normalized across all channels using the softmax function, yielding
\begin{align*}
     \boldsymbol{\alpha}_{gcn,i}^{(\ell)}  &=  
     \frac{\exp(\boldsymbol{e}_{gnc,i}^{(\ell)})}{\sum_{j=1}^{C_{gcn}} \exp(\boldsymbol{e}_{gcn,j}^{(\ell)})+\sum_{k=1}^{C_{sct}} \exp(\boldsymbol{e}_{sct,k}^{(\ell)})},
\end{align*}
with analogous $\boldsymbol{\alpha}_{sct,i}^{(\ell)}$. Note that the exponential function is applied elementwise here.
To obtain comparable weights when aggregating the $C_{gcn} + C_{sct}\eqqcolon C$ channels, we set
\begin{equation*}
     \boldsymbol{H}^{(\ell)} = C^{-1} \, \sigma\bigg( \sum_{j=1}^{C_{gcn}} \boldsymbol{\alpha}_{gcn,j}^{(\ell)} \odot \boldsymbol{\bar H}_{gcn,j}^{(\ell)}+\sum_{k=1}^{C_{sct}} \boldsymbol{\alpha}_{sct,k}^{(\ell)} \odot \boldsymbol{\bar H}_{sct,k}^{(\ell)} \bigg),
\end{equation*}
where $\sigma(.) = \operatorname{ReLU}(.)$ is used as nonlinearity here. 

%
\begin{table*}
\caption{Dataset characteristics \& comparison of node classification test accuracy. Datasets are ordered by increasing homophily.}\label{tab:data_accu}
\centering
\begin{tabular}{|c||c|c|c|c||
c|c|c||c|}
\hline
Dataset  & Classes & Nodes & Edges & Homophily & GCN & GAT & Sc-GCN & GSAN (ours) \\\hline
Texas & 5 & 183 &295 & 0.11 & 59.5 &58.4 & 60.3 & 60.5\\ \hline
Chameleon & 5 & 2,277 & 31,421 &0.23 & 28.2 & 42.9 & 51.2 & 61.2 \\ \hline
CoraFull & 70 & 19,793 &63,421 &0.57 & 62.2 & 51.9 & 62.5 & 64.3 \\ \hline
Wiki-CS & 10 & 11,701 &216,123 &0.65 &77.2 &77.7 & 78.1 & 78.6\\ \hline
Citeseer & 6 & 3,327 &4,676 &0.74 &70.3 &72.5 & 71.7 & 71.3\\ \hline
Pubmed & 3 & 19,717 &44,327 &0.80 &79.0 &79.0 & 79.4 & 79.8\\ \hline
Cora & 7 & 2,708 &5,276 &0.81 &81.5 &83.0 & 84.2 & 84.0\\ \hline
DBLP & 4 & 17,716 &52,867 &0.83 &59.3 &66.1 & 81.5 & 82.6\\ \hline
\end{tabular}
\end{table*}
\section{Scattering attention network}

\begin{figure}
    \centering
    \includegraphics[width=0.85\columnwidth]{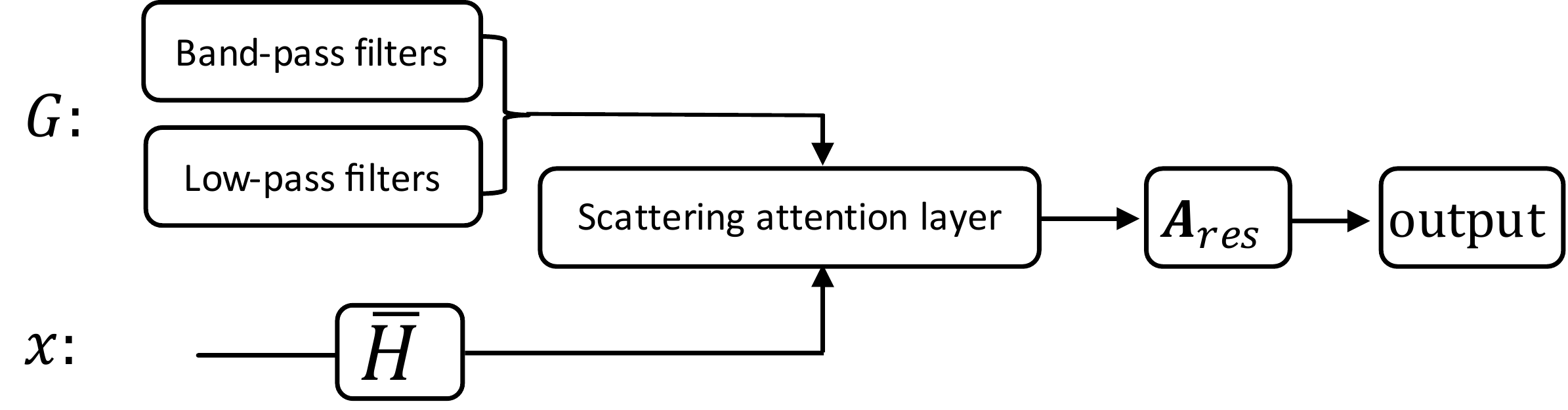}
    
    \caption{Illustration of the proposed network architecture.}
    \label{fig:architecture}
\end{figure}

\begin{figure}
\centering
\includegraphics[width=0.75\columnwidth]{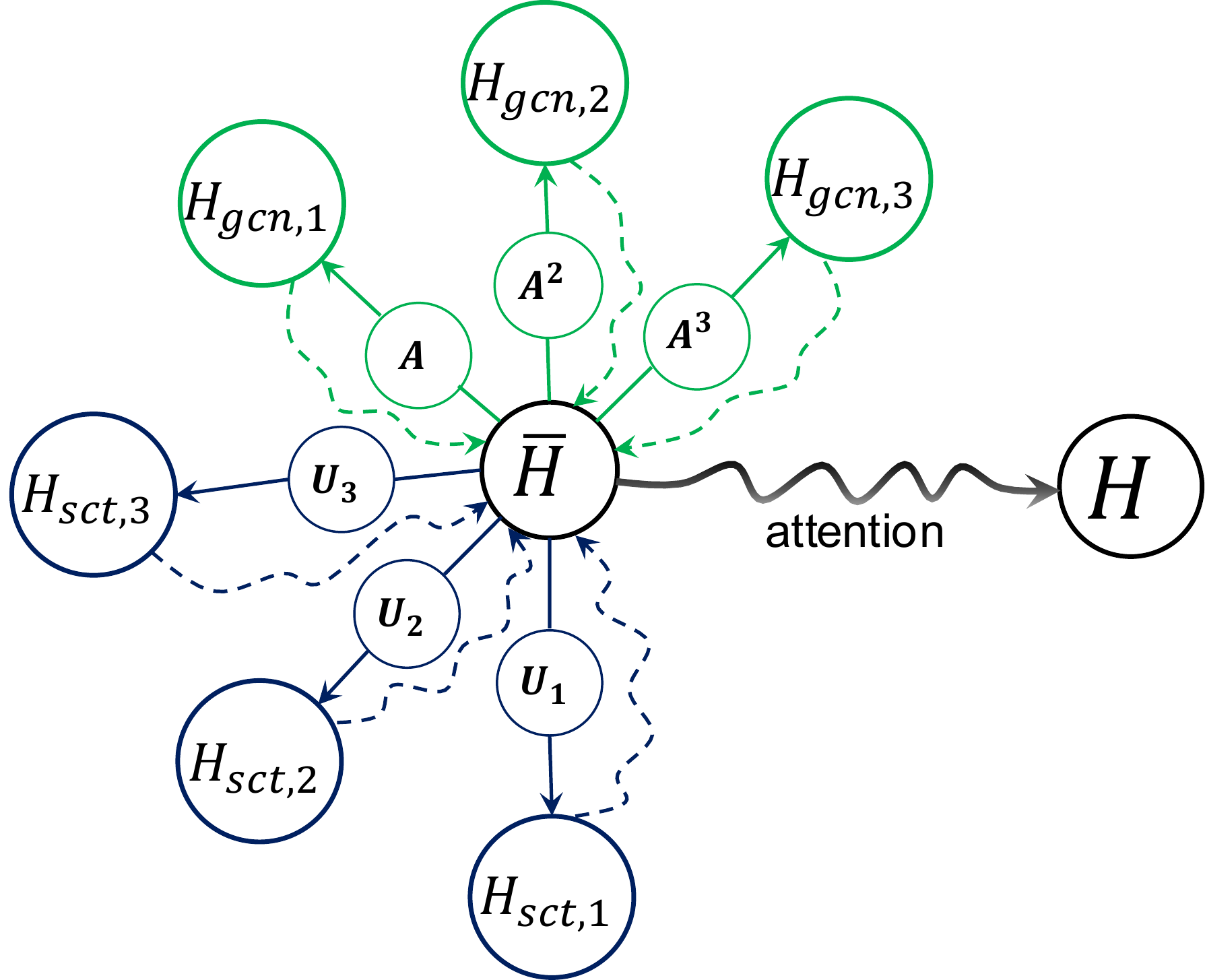}
\caption{Illustration of the proposed scattering attention layer. Attention weights are computed from a concatenation of the transformed layer input $\boldsymbol{\bar H}$ together with filtered signals that are first computed from it and then used to produce the layer output $\boldsymbol{H}$ via an attention-weighted linear combination.}
\label{fig:atten}
\end{figure}

The full network architecture in all examples shown here uses one scattering attention layer (as described in Sec.~\ref{sec:attentionlayer}), applied to the input node features, followed by a residual convolution layer (see Sec.~\ref{subsec_residual convolution}), which then produces the output via a fully connected layer. We note that deeper configurations are possible, especially when processing big graphs, but for simplicity, we focus on having a single architecture for the network. This is similar to the design choice utilized in GAT~\cite{velivckovic2018graph}. Figure~\ref{fig:architecture} illustrates this network structure and further technical details on each of its components are provided below.

\smallskip

\noindent\textbf{Attention layer configuration.} In this work, for simplicity, we set $C_{gcn} = C_{sct} = 3$, thus the attention layer combines three low-pass channels and three band-pass channels. The aggregation process of the attention layer is shown in Fig.~\ref{fig:atten}, where $\boldsymbol{U}_{1,2,3}$ represents three first-order scattering transformations with $\boldsymbol{U}_1 \boldsymbol{x} \coloneqq  \boldsymbol{\Psi}_1 x$,  $\boldsymbol{U}_2 \boldsymbol{x} \coloneqq \boldsymbol{\Psi}_2 x$ and $\boldsymbol{U}_3 \boldsymbol{x} \coloneqq \boldsymbol{\Psi}_3 x$. 

\smallskip

\noindent\textbf{Multihead attention.} Similar to other applications of attention mechanisms~\cite{velivckovic2018graph}, we use multi-head attention here for stabilizing the training, thus rewriting the output of the $\ell$-th layer (by a slight abuse of notation) as
\begin{equation}\label{eq:multi_head}
     \boldsymbol{H}^{(\ell)} \longleftarrow \Vert_{\gamma=1}^\Gamma  \boldsymbol{H}^{(\ell)}\left[\boldsymbol{\Theta}^{(\ell)}\mapsto \boldsymbol{\Theta}_{\gamma}^{(\ell)}; \boldsymbol{\alpha}^{(\ell)}\mapsto \boldsymbol{\alpha}_{\gamma}^{(\ell)}\right] ,
\end{equation}
combining $\Gamma$ attention heads, where $\Gamma$ is tuned as a hyperparameter of the network.

\smallskip

\noindent\textbf{Residual convolution.} To eliminate high frequency noise 
, the graph residual convolution (Sec.~\ref{subsec_residual convolution}) is applied to the output of the concatenated multi-head scattering scattering attention layer (Eq.~\ref{eq:multi_head}), with $\alpha$ tuned as a hyperparameter of the network.


\section{Results}
\label{sec:results}
To evaluate our geometric scattering attention network (GSAN), we apply it to semi-supervised node classification and compare its results on several benchmarks to two popular graph neural networks (namely GCN~\cite{kipf2016semi} and GAT~\cite{velivckovic2018graph}), as well as the original Sc-GCN, which does not utilize attention mechanisms and is instead tuned via extensive hyperparameter grid search. These methods are applied to eight benchmark datasets of varied sizes and homophily (i.e., average class similarity across edges), as shown in Tab.~\ref{tab:data_accu}. Texas and Chameleon are low-homophily datasets where nodes correspond to webpages and edges to links between them, with classes corresponding to webpage topic or monthly traffic (discretized into five levels), respectively~\cite{pei2019geom}. Wiki-CS is a recently proposed benchmark, where the nodes represent computer science articles and the edges represent the hyperlinks~\cite{mernyei2020wiki}.
The rest of the datasets are citation networks from different sources (i.e., Cora, Citeseer, Pubmed, DBLP), where nodes correspond to papers and edges to citations~\cite{yang2016revisiting,pan2016tri}. CoraFull is the larger version of the Cora dataset~\cite{bojchevski2018deep}.

All datasets are split into train, validation and test sets. The validation set is used for hyperparameter selection via grid search, including the number of heads $\Gamma$, the residual parameter $\alpha$ and the channel widths (i.e., number of neurons).
\begin{figure}[ht]
    \centering
    \includegraphics[width=\columnwidth]{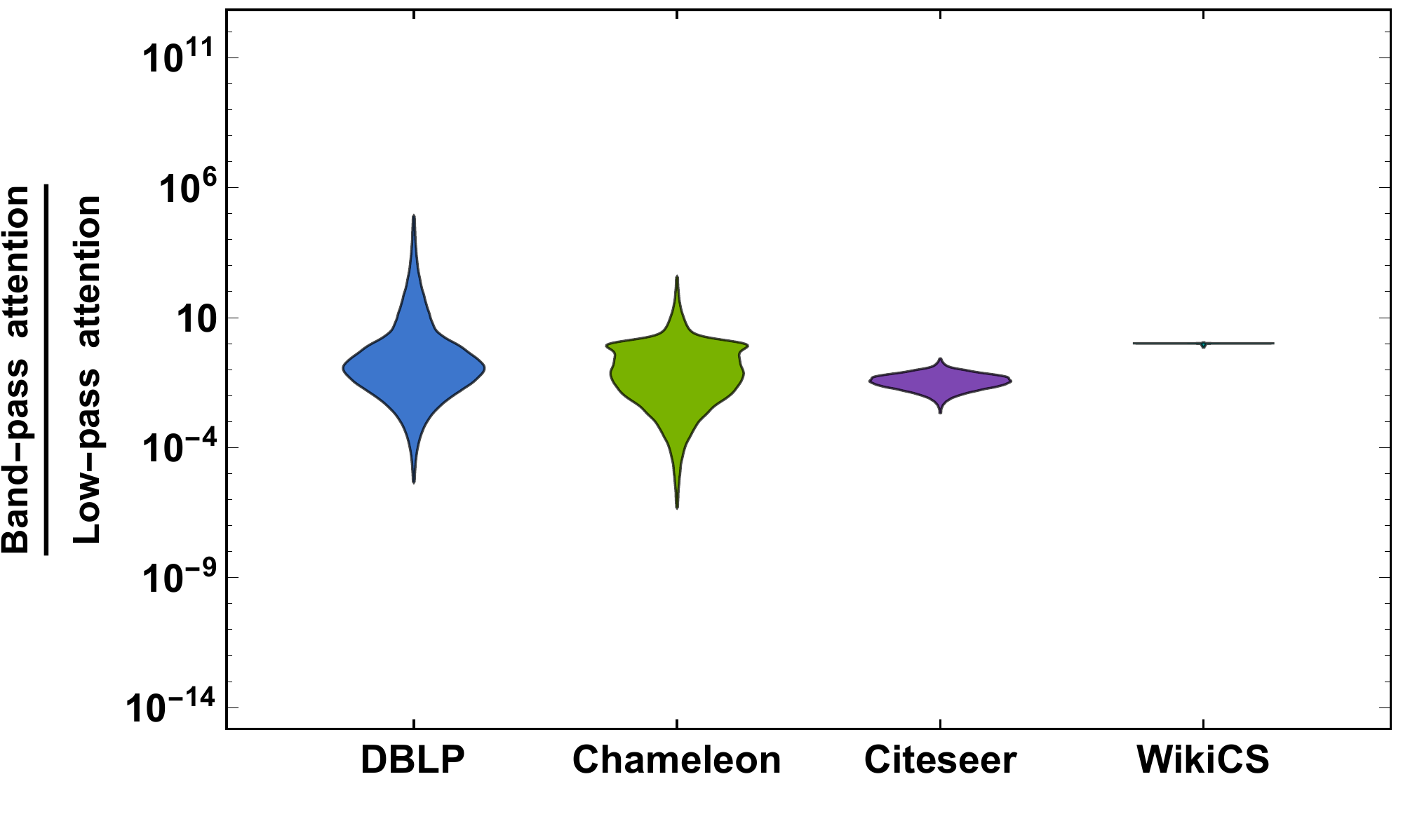}
    \vspace{-15pt}
    \caption{Distribution of attention ratio between band-pass (scattering) and low-pass (GCN) channels across nodes and heads in DBLP, Chameleon, Citeseer, and WikiCS.}
    \label{fig:attention-correlation}
\end{figure}

The results in Tab.~\ref{tab:data_accu} indicate that we improve upon previous methods, including the hybrid Sc-GCN that requires more intricate hyperparameter tuning to balance low-pass and band-pass channels~\cite{min2020scattering}. Since the attention weights $\boldsymbol{\alpha}$ are computed separately for every node (see Sec.~\ref{sec:attentionlayer}), they can also help understand the utilization of different channels in different regions of the graph. To demonstrate such analysis, we consider here the ratio between node-wise attention assigned to band-pass and low-pass channels. Over nodes and heads, we sum up the total attention $\sum_{i=1}^{C_{sct}}  \boldsymbol{1}_n^T\boldsymbol{\alpha}_{sct,i}$ in the three scattering channels $\boldsymbol{U}_{1,2,3}$, and $\sum_{i=1}^{C_{gcn}} \boldsymbol{1}_n^T\boldsymbol{\alpha}_{gcn,i}$ in the three GCN channels $\boldsymbol{A}^{1,2,3}$. Finally, we compute the band-pass vs.\ low-pass ratio $\frac{\sum_{i=1}^{C_{sct}}  \boldsymbol{1}_n^T\boldsymbol{\alpha}_{sct,i}}{\sum_{i=1}^{C_{gcn}}  \boldsymbol{1}_n^T\boldsymbol{\alpha}_{gcn,i}}$. 
Figure~\ref{fig:attention-correlation} demonstrates the additional insight provided by the distribution of these attention scores on four of the benchmark datasets. Wider spread, indicating highly varied channel utilization, is exhibited by DBLP and Chameleon where GSAN achieves significant improvement over GCN and GAT. Further, the improvement of GSAN over Sc-GCN on Chameleon highlights the importance of the node-wise feature selection in low-homophily settings. While Citeseer and Wiki-CS exhibit smaller spreads, the latter attributes more attention to band-pass channels, which we interpret as related to lower homophily.
\section{Conclusions}
The presented geometric scattering attention network (GSAN) introduces a new approach that leverages node-wise attention to incorporate both geometric scattering~\cite{gao2019geometric, gama2019diffusion, gama2019stability, zou2020graph} and GCN~\cite{kipf2016semi} channels to form a hybrid model, further advancing the recently proposed Sc-GCN~\cite{min2020scattering}. Beyond its efficacy in semi-supervised node classification, the distribution of its learned attention scores provides a promising tool to study the spectral composition of information extracted from node features. We expect this to enable future work to distill tractable notions of regularity on graphs to better understand and leverage their intrinsic structure in geometric deep learning, and to incorporate such attention mechanisms in spectral GNNs to both learn filter banks and perform node-wise selection of specific filters used in each local region of the graph.
\clearpage
\bibliographystyle{IEEEbib}
\bibliography{references}

\begin{thebibliography}{10}

\bibitem{bronstein2017geometric}
Michael~M. Bronstein, Joan Bruna, Yann LeCun, Arthur Szlam, and Pierre
  Vandergheynst,
\newblock ``Geometric deep learning: Going beyond {E}uclidean data,''
\newblock {\em IEEE Sig. Proc. Mag.}, vol. 34, no. 4, pp. 18--42, 2017.

\bibitem{kipf2016semi}
Thomas~N. Kipf and Max Welling,
\newblock ``Semi-supervised classification with graph convolutional networks,''
\newblock in {\em the 4th ICLR}, 2016.

\bibitem{velivckovic2018graph}
Petar Veli{\v{c}}kovi{\'c}, Guillem Cucurull, Arantxa Casanova, Adriana Romero,
  Pietro Li{\`o}, and Yoshua Bengio,
\newblock ``Graph attention networks,''
\newblock in {\em the 6th ICLR}, 2018.

\bibitem{hamilton2017inductive}
Will Hamilton, Zhitao Ying, and Jure Leskovec,
\newblock ``Inductive representation learning on large graphs,''
\newblock in {\em Advances in NeurIPS}, 2017, vol.~30, pp. 1024--1034.

\bibitem{bruna2013spectral}
Joan Bruna, Wojciech Zaremba, Arthur Szlam, and Yann LeCun,
\newblock ``Spectral networks and locally connected networks on graphs,''
\newblock arXiv:1312.6203, 2013.

\bibitem{defferrard2016convolutional}
Micha{\"e}l Defferrard, Xavier Bresson, and Pierre Vandergheynst,
\newblock ``Convolutional neural networks on graphs with fast localized
  spectral filtering,''
\newblock in {\em Advances in NeurIPS}, 2016, vol.~29.

\bibitem{nt2019revisiting}
Hoang NT and Takanori Maehara,
\newblock ``Revisiting graph neural networks: All we have is low-pass
  filters,''
\newblock arXiv:1905.09550, 2019.

\bibitem{dwivedi2020benchmarking}
Vijay~Prakash Dwivedi, Chaitanya~K Joshi, Thomas Laurent, Yoshua Bengio, and
  Xavier Bresson,
\newblock ``Benchmarking graph neural networks,''
\newblock arXiv:2003.00982, 2020.

\bibitem{li2018deeper}
Qimai Li, Zhichao Han, and Xiao-Ming Wu,
\newblock ``Deeper insights into graph convolutional networks for
  semi-supervised learning,''
\newblock in {\em Proc. of the 32nd AAAI Conf. on AI}, 2018.

\bibitem{oono2019graph}
Kenta Oono and Taiji Suzuki,
\newblock ``Graph neural networks exponentially lose expressive power for node
  classification,''
\newblock in {\em the 7th ICLR}, 2019.

\bibitem{gao2019geometric}
Feng Gao, Guy Wolf, and Matthew Hirn,
\newblock ``Geometric scattering for graph data analysis,''
\newblock in {\em Proceedings of the 36th ICML}, 2019, pp. 2122--2131.

\bibitem{gama2019diffusion}
Fernando Gama, Alejandro Ribeiro, and Joan Bruna,
\newblock ``Diffusion scattering transforms on graphs,''
\newblock in {\em the 7th ICLR}, 2019.

\bibitem{gama2019stability}
Fernando Gama, Alejandro Ribeiro, and Joan Bruna,
\newblock ``Stability of graph scattering transforms,''
\newblock in {\em Advances in NeurIPS}, 2019, vol.~32, pp. 8038--8048.

\bibitem{zou2020graph}
Dongmian Zou and Gilad Lerman,
\newblock ``Graph convolutional neural networks via scattering,''
\newblock {\em Applied and Computational Harmonic Analysis}, vol. 49, no. 3,
  pp. 1046--1074, 2020.

\bibitem{mallat2012group}
St{\'e}phane Mallat,
\newblock ``Group invariant scattering,''
\newblock {\em Communications on Pure and Applied Mathematics}, vol. 65, no.
  10, pp. 1331--1398, 2012.

\bibitem{bruna:invariantScatConvNet2013}
Joan Bruna and St\'{e}phane Mallat,
\newblock ``Invariant scattering convolution networks,''
\newblock {\em IEEE Trans. on Patt. Anal. and Mach. Intel.}, vol. 35, no. 8,
  pp. 1872--1886, August 2013.

\bibitem{anden:deepScatSpectrum2014}
Joakim And\'{e}n and St\'{e}phane Mallat,
\newblock ``Deep scattering spectrum,''
\newblock {\em IEEE Trans. on Sig. Proc.}, vol. 62, no. 16, pp. 4114--4128,
  August 2014.

\bibitem{min2020scattering}
Yimeng Min, Frederik Wenkel, and Guy Wolf,
\newblock ``Scattering gcn: Overcoming oversmoothness in graph convolutional
  networks,''
\newblock {\em Advances in NeurIPS}, vol. 33, 2020.

\bibitem{bianchi2021graph}
Filippo~Maria Bianchi, Daniele Grattarola, Lorenzo Livi, and Cesare Alippi,
\newblock ``Graph neural networks with convolutional arma filters,''
\newblock {\em IEEE Trans. on Patt. Anal. and Mach. Intel.}, 2021.

\bibitem{mallat:rotoScat2013}
Laurent Sifre and St\'{e}phane Mallat,
\newblock ``Rotation, scaling and deformation invariant scattering for texture
  discrimination,''
\newblock in {\em the 2013 CVPR}, June 2013.

\bibitem{lostanlen:scatPitchSpiral2015}
Vincent Lostanlen and St\'{e}phane Mallat,
\newblock ``Wavelet scattering on the pitch spiral,''
\newblock in {\em Proc. of the 18th International Conference on Digital Audio
  Effects}, 2015, pp. 429--432.

\bibitem{coifman:diffWavelets2006}
R.R. Coifman and M.~Maggioni,
\newblock ``Diffusion wavelets,''
\newblock {\em Applied and Computational Harmonic Analysis}, vol. 21, no. 1,
  pp. 53--94, 2006.

\bibitem{pei2019geom}
Hongbin Pei, Bingzhe Wei, Kevin Chen-Chuan Chang, Yu~Lei, and Bo~Yang,
\newblock ``Geom-gcn: Geometric graph convolutional networks,''
\newblock in {\em the 7th ICLR}, 2019.

\bibitem{mernyei2020wiki}
P{\'e}ter Mernyei and C{\u{a}}t{\u{a}}lina Cangea,
\newblock ``Wiki-cs: A wikipedia-based benchmark for graph neural networks,''
\newblock arXiv:2007.02901, 2020.

\bibitem{yang2016revisiting}
Zhilin Yang, William Cohen, and Ruslan Salakhudinov,
\newblock ``Revisiting semi-supervised learning with graph embeddings,''
\newblock in {\em Proceedings of the 33rd ICML}, 2016, vol.~48 of {\em PMLR},
  pp. 40--48.

\bibitem{pan2016tri}
Shirui Pan, Jia Wu, Xingquan Zhu, Chengqi Zhang, and Yang Wang,
\newblock ``Tri-party deep network representation,''
\newblock in {\em Proc. of the 25th IJCAI}, 2016, pp. 1895--1901.

\bibitem{bojchevski2018deep}
Aleksandar Bojchevski and Stephan G{\"u}nnemann,
\newblock ``Deep gaussian embedding of graphs: Unsupervised inductive learning
  via ranking,''
\newblock in {\em the 6th ICLR}, 2018.

\end{thebibliography}
\end{document}